\documentclass[11pt]{article}

\usepackage[final]{acl}

\usepackage{times}
\usepackage{latexsym}

\usepackage[T1]{fontenc}

\usepackage[utf8]{inputenc}

\usepackage{microtype}

\usepackage{inconsolata}
\usepackage{amssymb} 
\usepackage[table,xcdraw]{xcolor} 
\usepackage{arydshln} 
\usepackage{makecell} 
\usepackage{multirow} 
\usepackage{graphicx} 

\usepackage{hhline}
\usepackage{pifont}
\usepackage{colortbl}
\usepackage{arydshln}

\newcommand{\cmark}{\textcolor[HTML]{4CAF50}{\ding{51}}}
\newcommand{\xmark}{\textcolor[HTML]{E57373}{\ding{55}}}

\usepackage{xspace}
\usepackage{hyperref}
\usepackage{amssymb}
\usepackage{amsmath}
\usepackage{nccmath}
\usepackage{booktabs}
\usepackage{multirow}
\usepackage[table]{xcolor} 
\usepackage{float} 
\usepackage{cuted}
\newcommand{\boldparagraph}[1]{\noindent\textbf{#1}\ }
\definecolor{bestbg}{RGB}{198,234,212}   
\definecolor{secondbg}{RGB}{231,242,255} 
\definecolor{secondbg2}{RGB}{255, 218, 224}
\definecolor{mygreen}{HTML}{00AA00}

\newcommand{\huggingface}{\raisebox{-1.5pt}{\includegraphics[height=1.05em]{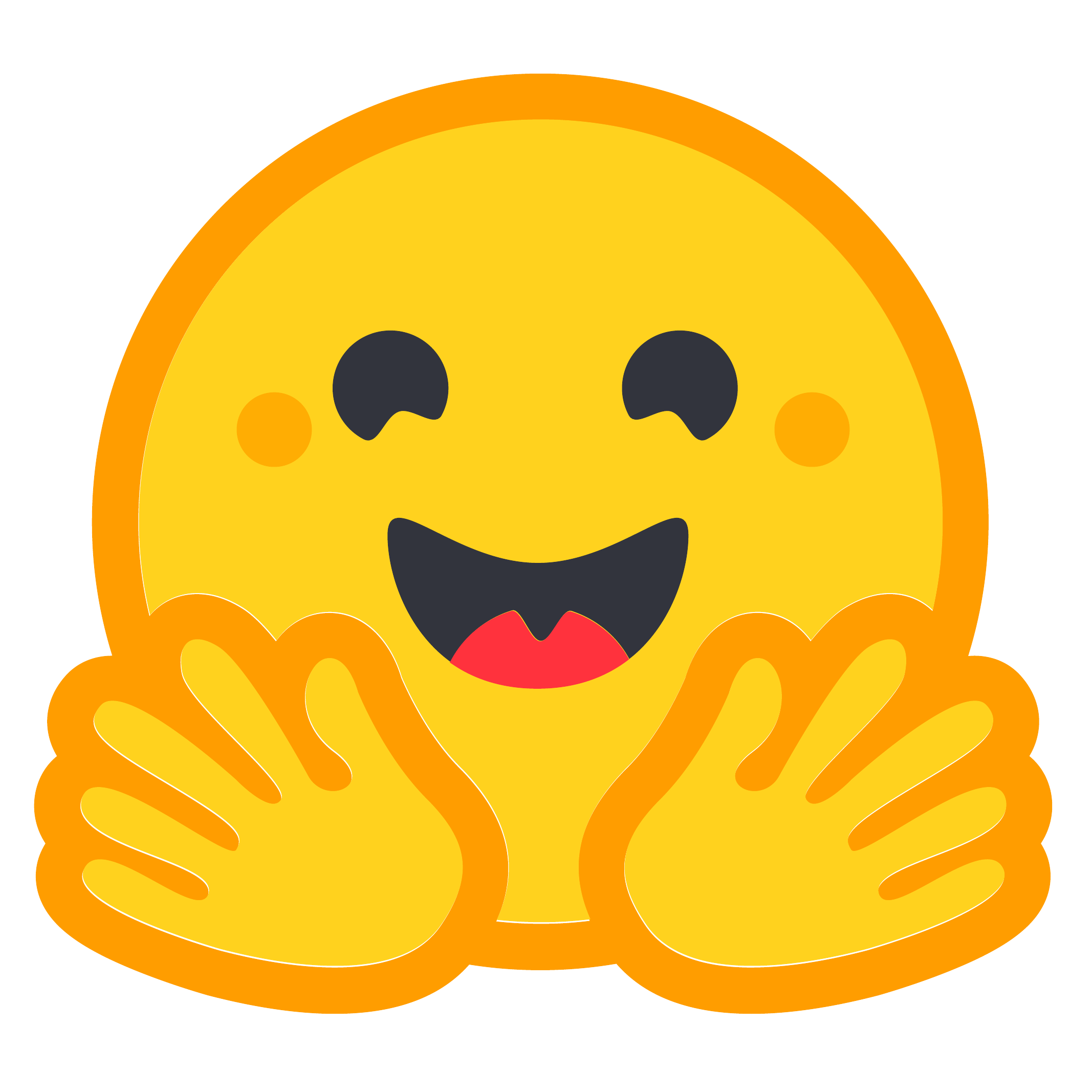}}\xspace}

\newcommand{\github}{\raisebox{-1.5pt}{\includegraphics[height=1.05em]{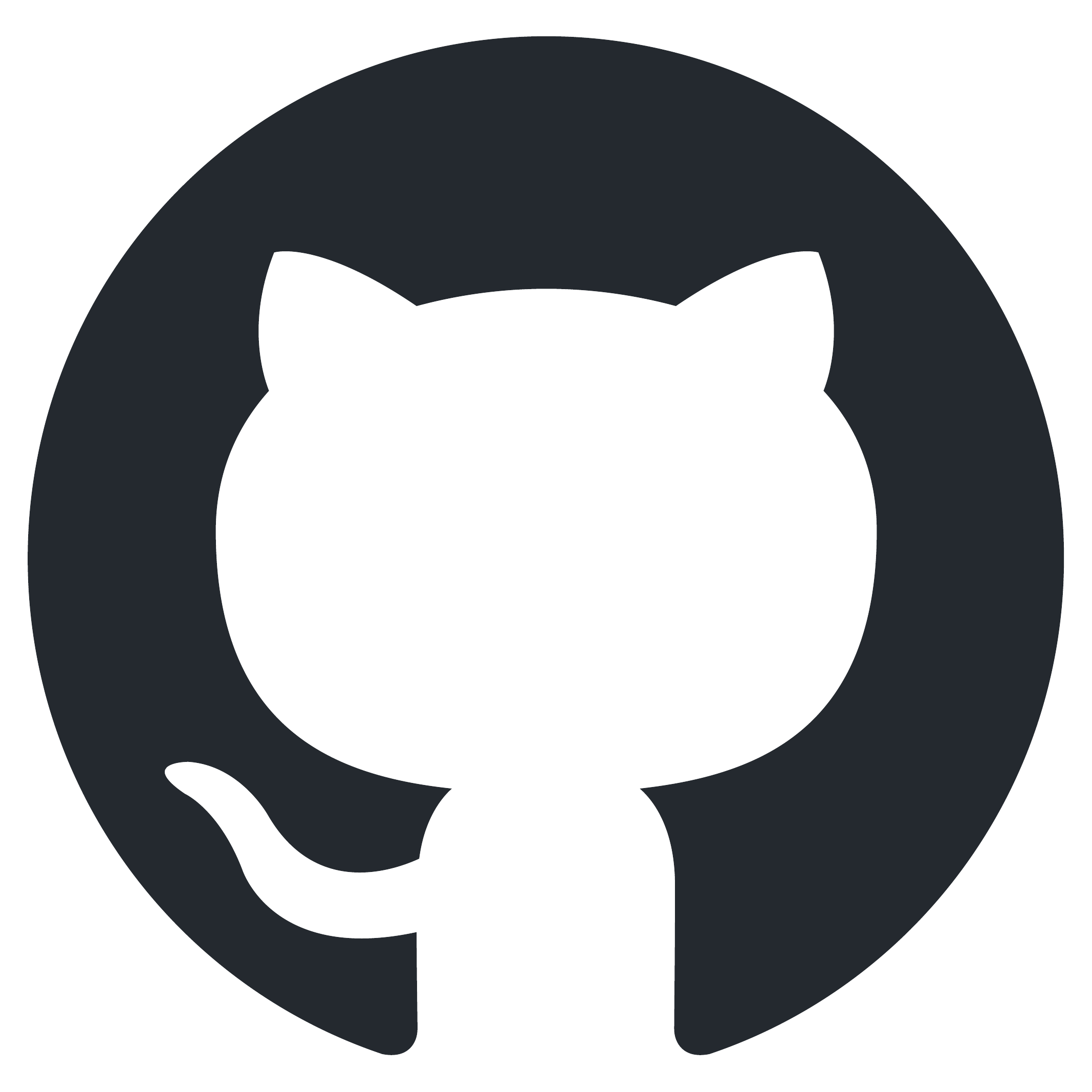}}\xspace}
\newcommand{\project}{\raisebox{0pt}{\includegraphics[height=1.0em]{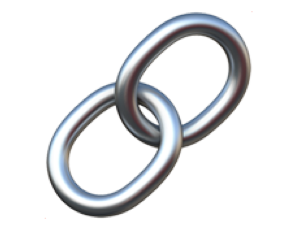}}\xspace}

\title{Beyond Global Scalars: Synergizing Token-Level Statistics and\\[0.3em] Deep Semantics for Adversarial AIGC Text Detection}

\author{
Peiming Li$^{1,2}$ , Yifan Wang$^{1}$ , Zhiyuan Hu$^{1,2}$ ,  Shiyu Li$^{1}$ , \\
\textbf{Zheng Wei$^{1}$$^{,}$\footnotemark[1]}, \textbf{Yang Tang$^{1}$$^{,}$\thanks{Corresponding authors.}$^{,}$\thanks{Project Lead.}} \\
$^1$Tencent BAC \\
$^2$School of Electronic and Computer Engineering, Peking University \\
\small{\texttt{{hemingwei@tencent.com, ethanntang@tencent.com}}}
\\[0.4em]
{\small
\project\!\href{https://tencentbac.github.io/NeuroStat/}{Homepage} \;\textbar\;
\github\!\href{https://github.com/TencentBAC/NeuroStat}{Code} \;\textbar\;
\huggingface\!\href{https://huggingface.co/collections/TencentBAC/neurostat}{MOSAIC Dataset}
}
}

\begin{document}
\maketitle
\begin{abstract}
The rapid evolution of large language models necessitates robust machine-generated text detection. 
Existing paradigms typically follow two isolated tracks. 
Training-free methods rely on global statistical scalars such as perplexity, while training-based methods utilize semantic hidden states. 
Both approaches exhibit fundamental vulnerabilities in adversarial scenarios. 
Global scalars act as lossy compressions that obscure local probabilistic burstiness in interleaved texts, whereas pure semantic models overfit to specific fingerprints and remain susceptible to spoofing. 
To expose these flaws, we introduce \textbf{MOSAIC}, a comprehensive adversarial benchmark comprising 16000 samples across a full-granularity attack spectrum. 
To address these challenges, we propose \textbf{NeuroStat}, an end-to-end framework bridging the statistical and semantic gap. 
NeuroStat captures uncompressed token-level probabilistic logits alongside deep semantic hidden states from a single causal language model backbone. 
We fuse these heterogeneous signals through Macro-State Residual Modulation, which adaptively calibrates local convolutional features using global uncertainty indicators. 
Orthogonal and contrastive losses further ensure the learning of complementary representations. 
Extensive experiments demonstrate that NeuroStat maintains exceptional robustness on MOSAIC compared to the severe degradation of state-of-the-art methods, establishing a new standard for adversarial text detection. Code and the MOSAIC benchmark are available at \url{https://github.com/TencentBAC/NeuroStat}.
\end{abstract}

\begin{figure*}[t]
   \centering
   \includegraphics[width=1.0\linewidth]{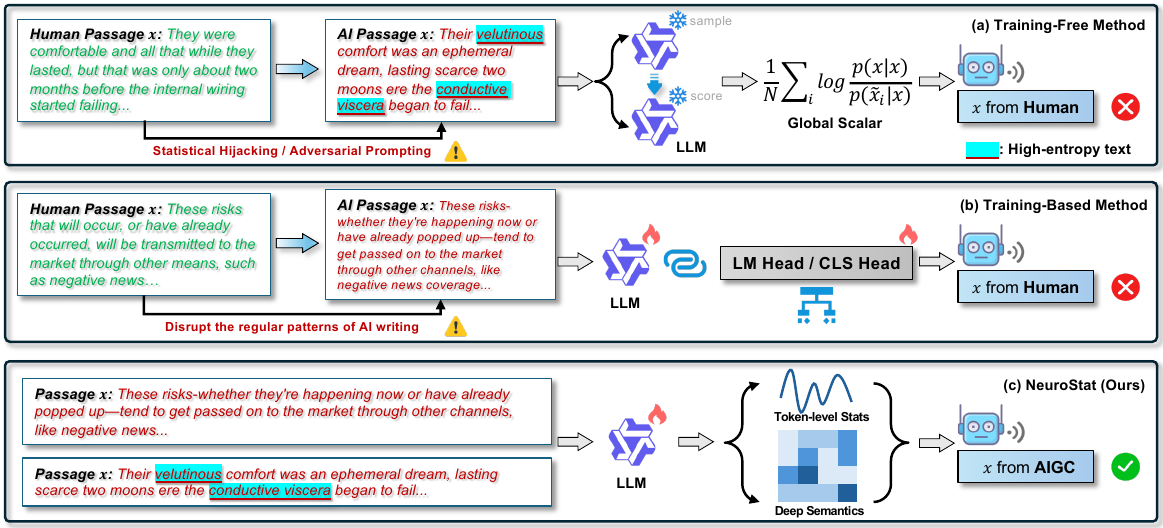}
   \caption{\textbf{Comparison of detection paradigms under adversarial attacks.} Existing methods fail due to their reliance on either (a) easily hijacked global statistical scalars (Training-Free) or (b) overfitted semantic patterns that can be disrupted by style manipulation (Training-Based). (c) NeuroStat robustly detects adversarial AIGC by synergizing uncompressed token-level statistical trajectories and deep semantic representations.}
   \vspace{-2pt}
   \label{fig:intro}
\end{figure*}

\section{Introduction}

The proliferation of advanced Large Language Models (LLMs)~\cite{brown2020language, touvron2023llama, openai2023gpt4,glm5team2026glm5vibecodingagentic,kimiteam2026kimik25visualagentic} has blurred the boundary between human-written and machine-generated text~\cite{clark2021all}. While this technological leap empowers productivity, it exacerbates the spread of misinformation~\cite{zellers2019defending}. Consequently, Machine-Generated Text Detection (MGTD) has emerged as a critical research frontier.

Current MGTD methodologies predominantly follow two isolated paradigms, Training-Free (TF) methods that calculate global statistical scalars (e.g., perplexity)~\cite{mitchell2023detectgpt, bao2023fast}, and Training-Based (TB) methods that fine-tune black-box encoders to capture semantic styles~\cite{solaiman2019release, verma2023ghostbuster}. Recent hybrid strategies~\cite{chen2025imitate, fu2025detectanyllm} attempt to optimize scoring models but still rely on lossy global scalars. In real-world applications, malicious users rarely generate entire documents directly. Instead, they employ Interleaved Human-AI Writing or Statistical Hijacking~\cite{sadasivan2023can, krishna2023paraphrasing}. As illustrated in Fig.~\ref{fig:intro}, existing paradigms are easily bypassed, TF methods average out local probability bursts and are spoofed by high-entropy instructions (Fig.~\ref{fig:intro}a), while TB methods overfit to specific semantic fingerprints, leaving them susceptible to style manipulation (Fig.~\ref{fig:intro}b).

To systematically diagnose these vulnerabilities, we introduce \textbf{MOSAIC} (\textbf{M}ultidimensional \textbf{O}bfuscation and \textbf{S}poofing \textbf{A}dversarial \textbf{I}nterleaved \textbf{C}orpus), the most comprehensive adversarial MGTD benchmark to date. Unlike existing datasets~\cite{guo2023close, dugan2024raid}, MOSAIC constructs a full-granularity attack spectrum encompassing 8 categories and 36 sub-methods, employing a cross-allocation mechanism across six frontier LLMs to mitigate model fingerprint bias.

To overcome the inherent flaws exposed by MOSAIC, we propose \textbf{NeuroStat} (\textbf{Neu}ral Embeddings and \textbf{R}epresentations \textbf{O}rthogonalized with \textbf{Sta}tistical \textbf{T}rajectories), a novel end-to-end framework that bridges the TF-TB methods' gap. We argue that TF and TB provide orthogonal but complementary views, TF captures the probabilistic mechanics of decoding, while TB captures contextual semantics. NeuroStat extracts both token-level statistical trajectories and deep semantic representations from a single CausalLM backbone. 

Crucially, instead of compressing probabilities into a vulnerable scalar, NeuroStat preserves the full sequential trajectory. To fuse these heterogeneous signals, we design a Macro-State Residual Modulation (MSRM) mechanism. MSRM utilizes global uncertainty indicators to dynamically amplify local probability anomalies, effectively capturing hidden AI bursts within fluent human contexts. Trained with orthogonal and contrastive objectives, NeuroStat learns discriminative and non-redundant representations, maintaining exceptional robustness against adversarial attacks (Fig.~\ref{fig:intro}c).

Our contributions are summarized as follows:
\begin{itemize}
    \item \textbf{A Paradigm Shift in Detection Architecture.} We propose NeuroStat, the first framework to achieve end-to-end fusion of probabilistic trajectories and semantic artifacts. By replacing lossy global scalars with MSRM, it overcomes dilution and spoofing flaws.
    \item \textbf{A Full-Spectrum Adversarial Benchmark:} We construct MOSAIC, a benchmark featuring an unprecedented 8$\times$36 attack matrix and multi-model cross-generation, establishing a new standard for MGTD robustness.
    \item \textbf{State-of-the-Art Robustness.} Extensive experiments demonstrate that NeuroStat significantly outperforms existing baselines across diverse domains, models, and attack granularities, maintaining high accuracy even in extreme adversarial scenarios.
\end{itemize}

\section{Related Work}
\label{sec:related_work}

\boldparagraph{Training-Free (TF) Detection Methods.}
TF methods distinguish machine-generated text by leveraging LLMs' inherent probabilistic mechanics via zero-shot metrics, including log-likelihood, entropy~\cite{lavergne2008detecting, solaiman2019release, gehrmann2019gltr}, probability curvature~\cite{mitchell2023detectgpt, bao2023fast}, log-rank~\cite{su2023detectllm}, n-gram divergence~\cite{yang2023dna}, cross-perplexity~\cite{hans2024spotting}, and distribution recovery~\cite{bao2024glimpse}. However, these methods universally compress sequential token-level dynamics into a lossy global scalar, making them highly vulnerable to dilution in interleaved human-AI texts and statistical hijacking.

\boldparagraph{Training-Based (TB) Detection Methods.}
TB methods treat detection as a supervised task to capture semantic artifacts, employing encoder fine-tuning~\cite{liu2019roberta, zellers2019defending, automatic}, adversarial training~\cite{hu2023radar, verma2023ghostbuster, wu2023llmdet}, or alignment-based optimization~\cite{chen2025imitate, fu2025detectanyllm}. Despite near-perfect in-domain accuracy, TB methods inherently overfit to specific semantic fingerprints. Diverging from paradigms relying on isolated global scalars or black-box semantics, NeuroStat synergizes uncompressed probabilistic trajectories with semantic representations.

\boldparagraph{Detection Benchmarks and Adversarial Robustness.}
Early detection benchmarks focused on basic generation~\cite{uchendu2021turingbench, guo2023close} before expanding to multi-domain settings~\cite{wang2023m4, macko2023multitude, li2024mage}. Concurrently, studies reveal detectors are easily bypassed by paraphrasing, homoglyphs, and prompt engineering~\cite{sadasivan2023can, krishna2023paraphrasing, gagiano2021robustness, lu2023large}. While RAID~\cite{dugan2024raid} standardized robustness evaluation with 11 attacks, existing benchmarks still suffer from limited attack surfaces and model bias. To our knowledge, MOSAIC is the first to provide a full-granularity taxonomy of 36 attacks coupled with multi-model cross-generation.

\begin{figure*}[htbp]
   \centering
   \includegraphics[width=1.0\linewidth]{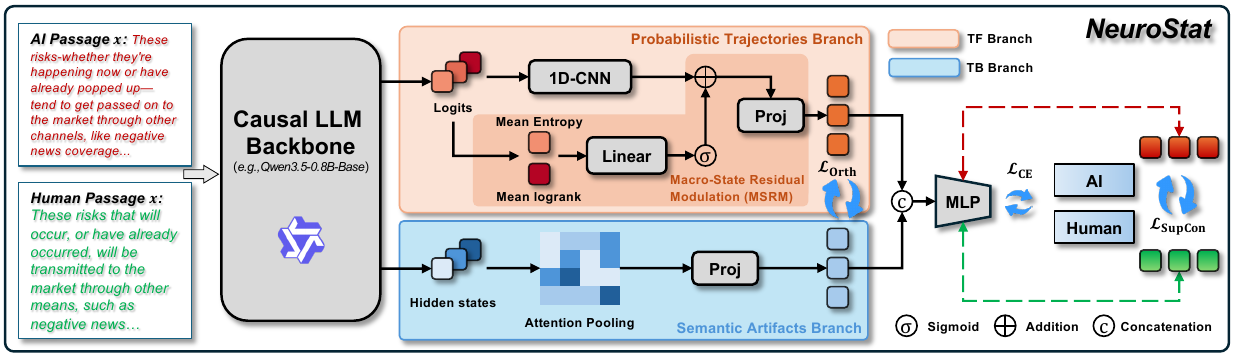}
   \caption{\textbf{The overall architecture of NeuroStat.} A CausalLM backbone simultaneously extracts token logits and hidden states. The \textbf{Probabilistic Trajectories Branch} (top) captures local statistical patterns, which are dynamically calibrated by the \textbf{MSRM} mechanism using global uncertainty indicators. Concurrently, the \textbf{Semantic Artifacts Branch} (bottom) extracts contextual styles via attention pooling. The fused representations are jointly optimized with $\mathcal{L}_{\mathrm{Orth}}$ and $\mathcal{L}_{\mathrm{SupCon}}$ objectives to ensure complementary and robust feature learning.}
   \vspace{-2pt}
   \label{fig:pipeline}
\end{figure*}

\section{Methodology}
\label{sec:method}

We present \textbf{NeuroStat}, an end-to-end dual-branch framework that bridges the Training-Free (TF) and Training-Based (TB) paradigms for robust machine-generated text detection (illustrated in Fig.~\ref{fig:pipeline}). The core philosophy of NeuroStat is to decouple the detection process into two orthogonal perspectives: the probabilistic mechanics of the decoding process (TF) and the contextual semantic artifacts (TB). Given an input token sequence $\mathbf{x} = (x_1, \ldots, x_T)$ of length $T$, a single forward pass through the backbone $\mathcal{M}$ yields the full vocabulary logit matrix $\mathbf{L} \in \mathbb{R}^{T \times |\mathcal{V}|}$ and the last-layer hidden states $\mathbf{H} \in \mathbb{R}^{T \times d}$, where $|\mathcal{V}|$ is the vocabulary size and $d$ is the hidden dimension. These heterogeneous signals are routed to their respective branches, enabling synergistic feature extraction without requiring multiple model invocations.

\subsection{Probabilistic Trajectories Branch (TF)}
\label{sec:tf_branch}
Existing TF methods typically compress token probabilities into a lossy global scalar (e.g., perplexity), which inadvertently smooths out local anomalies. Instead, our TF branch is designed to preserve the complete sequential structure of the probability landscape. Specifically, we extract the shifted log-probability trajectory $\boldsymbol{\ell} = (\ell_1, \ldots, \ell_{T-1})$ of the actual tokens, where:
\begin{equation}
    \ell_t = \log \big( \text{softmax}(\mathbf{L}_{t}) \big)_{x_{t+1}}.
\end{equation}
Here, $\mathbf{L}_t$ denotes the logit vector at step $t$, and $x_{t+1}$ is the actual next token in the sequence.

Unlike machine-generated texts that maintain consistently high probabilities, human-authored content inherently manifests pronounced variance in $\boldsymbol{\ell}$ (i.e., local burstiness). To capture these localized dynamics and identify abrupt likelihood decays indicative of human-AI boundaries, we process $\boldsymbol{\ell}$ through a hierarchical 1D-CNN:
\begin{equation}
    \mathbf{c} = f_{\text{CNN}}(\boldsymbol{\ell}) \in \mathbb{R}^{d_c}.
\end{equation}
This architecture functions as a localized $n$-gram anomaly detector, effectively extracting multi-scale probabilistic deviations while preserving strict positional awareness.

\boldparagraph{Macro-State Residual Modulation (MSRM).}
While the 1D-CNN effectively extracts localized dynamics, it lacks global contextual awareness. An abrupt probability decay could signify a naturally rare word in human writing or a hallucination boundary in AI generation. To disambiguate such signals, MSRM conditions the CNN features on two macro-state uncertainty indicators: mean entropy $\bar{E}$ and mean log-rank $\bar{R}$:
\begin{align}
    \bar{E} &= \frac{1}{T\!-\!1}\sum_{t=1}^{T-1} \mathcal{H}\big(\text{softmax}(\mathbf{L}_t)\big), \label{eq:mean_entropy} \\
    \bar{R} &= \frac{1}{T\!-\!1}\sum_{t=1}^{T-1} \log \text{rank}_t(x_{t+1}), \label{eq:mean_logrank}
\end{align}
where $\mathcal{H}(\cdot)$ denotes the Shannon entropy, and $\text{rank}_t(x_{t+1})$ is the descending rank of the target token at step $t$. These scalars, concatenated as $\mathbf{s} = [\bar{E}; \bar{R}] \in \mathbb{R}^{2}$, quantify the model's sequence-level confidence. This design formalizes an asymmetry long exploited by training-free detectors, where global confidence and local rank or probability deviations jointly determine separability rather than either signal alone~\cite{gehrmann2019gltr, mitchell2023detectgpt, hans2024spotting}. They modulate the CNN features via a residual gating mechanism:
\begin{equation}
    \tilde{\mathbf{c}} = \mathbf{c} \odot (1 + \sigma(\mathbf{W}_g \mathbf{s})),
    \label{eq:msrm}
\end{equation}
where $\mathbf{W}_g \in \mathbb{R}^{d_c \times 2}$ is a learnable projection and $\sigma$ is the sigmoid function. Conceptually, when the global entropy is low, reflecting the high overall confidence typical of AI generation, MSRM acts as an adaptive amplifier for localized anomaly signals. The modulated features are subsequently projected to yield the final TF representation $\mathbf{z}_{\text{tf}} \in \mathbb{R}^{d_z}$. We empirically verify this design by correlating the learned gate activation with the global entropy $\bar{E}$ on held-out samples, obtaining a Pearson correlation of $-0.62$. This confirms that the network learns to amplify local features precisely when global confidence is high, without this behavior being explicitly imposed.

\subsection{Semantic Artifacts Branch (TB)}
\label{sec:tb_branch}
While the TF branch focuses on how the text was generated, the TB branch focuses on what was generated. Machine-generated texts often contain specific stylistic markers, repetitive phrasing patterns, or rigid discourse structures (e.g., ``delve into'', ``in conclusion''). To capture these high-level semantic artifacts from the hidden states $\mathbf{H}$, we employ a learnable attention pooling mechanism:
\begin{equation}
    \alpha_t = \frac{\exp(\mathbf{w}_a^\top \mathbf{h}_t)}{\sum_{j=1}^{T} \exp(\mathbf{w}_a^\top \mathbf{h}_{j})}, \quad \mathbf{h}_{\text{pool}} = \sum_{t=1}^{T} \alpha_t \, \mathbf{h}_t,
\end{equation}
where $\mathbf{w}_a \in \mathbb{R}^{d}$ is a learnable query vector. Unlike relying solely on the last token, this attention mechanism enables the model to dynamically assign higher weights to the most semantically discriminative tokens, regardless of sequence length. The pooled representation is then projected to yield the TB representation $\mathbf{z}_{\text{tb}} \in \mathbb{R}^{d_z}$.

\subsection{Training Objectives}
\label{sec:fusion_loss}
The representations from both branches are concatenated and passed through a Multi-Layer Perceptron (MLP) to produce the binary classification logits $\hat{\mathbf{y}} = \text{MLP}([\mathbf{z}_{\text{tf}} ; \mathbf{z}_{\text{tb}}])$. To ensure that NeuroStat learns robust and non-redundant features, it is optimized with three complementary objectives:

\boldparagraph{Cross-Entropy Loss ($\mathcal{L}_{CE}$).} The primary classification loss optimized over the predicted probability of the AIGC class.

\boldparagraph{Supervised Contrastive Loss ($\mathcal{L}_{\mathrm{SupCon}}$).} To enhance the model's robustness against out-of-distribution LLMs, we apply $\mathcal{L}_{\mathrm{SupCon}}$ to the $\ell_2$-normalized fused vector $\mathbf{e}_i$:
\begin{equation}
\begin{aligned}
\mathcal{L}_{\mathrm{SupCon}}
&= \frac{-1}{B}\sum_{i=1}^{B} \frac{1}{|P(i)|}\sum_{j \in P(i)}
\log \\
&\quad \frac{\exp(\mathbf{e}_i \cdot \mathbf{e}_j / \tau)}
{\sum_{k \neq i} \exp(\mathbf{e}_i \cdot \mathbf{e}_k / \tau)},
\end{aligned}
\end{equation}
where $P(i)$ contains same-class samples in batch $B$, and $\tau$ is the temperature. This pulls same-class representations together while pushing apart different classes, improving robustness against distributional shifts.

\boldparagraph{Orthogonal Penalty ($\mathcal{L}_{\mathrm{Orth}}$).} A critical challenge in dual-branch architectures is feature collapse, where one branch dominates or both learn redundant information. Assuming both branches are projected to the same dimension $d_z$, we force the TF and TB branches to span orthogonal subspaces by minimizing their squared cosine similarity:
\begin{equation}
    \mathcal{L}_{\mathrm{Orth}} = \frac{1}{B}\sum_{i=1}^{B} \left( \frac{\mathbf{z}_{\text{tf}}^i \cdot \mathbf{z}_{\text{tb}}^i}{\|\mathbf{z}_{\text{tf}}^i\| \|\mathbf{z}_{\text{tb}}^i\|} \right)^2.
\end{equation}
This strict regularization ensures that the semantic branch does not redundantly encode statistical patterns, and vice versa. Measured on held-out samples, the mean squared cosine similarity between the two branches drops from 0.358 without $\mathcal{L}_{\mathrm{Orth}}$ to 0.018 with $\mathcal{L}_{\mathrm{Orth}}$, closely approaching the 0.016 floor observed for independent random unit vectors of the same dimensionality. This indicates that the two branches become effectively independent rather than merely less correlated.

\boldparagraph{Total Objective.} The overall training objective is formulated as:
\begin{equation}
    \mathcal{L} = \mathcal{L}_{CE} + \lambda_{\mathrm{SupCon}} \mathcal{L}_{\mathrm{SupCon}} + \lambda_{\mathrm{Orth}} \mathcal{L}_{\mathrm{Orth}},
\end{equation}
where $\lambda_{\mathrm{SupCon}}$ and $\lambda_{\mathrm{Orth}}$ are hyperparameters balancing the auxiliary losses.

\begin{table*}[htbp]
\centering
\small
\renewcommand{\arraystretch}{1}
\setlength{\tabcolsep}{2pt}
\resizebox{\textwidth}{!}{%
\begin{tabular}{l cc ccc cccccccc cc}
\Xhline{1.2pt}
\\[-8pt]
\multirow{2}{*}{\textbf{Benchmark}} 
  & \multicolumn{2}{c}{\textbf{Data Statistics}} 
  & \multicolumn{3}{c}{\textbf{MGT Tasks}} 
  & \multicolumn{8}{c}{\textbf{Adversarial Attack Coverage}} 
  & \multicolumn{2}{c}{\textbf{Summary}} \\
\cmidrule(lr){2-3} \cmidrule(lr){4-6} \cmidrule(lr){7-14} \cmidrule(lr){15-16}
  & \textbf{Size} & \textbf{\#Dom.}
  & \textbf{Gen.} & \textbf{Pol.} & \textbf{Rew.}
  & \makecell[c]{\footnotesize\textbf{Inter-}\\\footnotesize\textbf{leaved}}
  & \makecell[c]{\footnotesize\textbf{Stat.}\\\footnotesize\textbf{Hijack}}
  & \makecell[c]{\footnotesize\textbf{Trans-}\\\footnotesize\textbf{lation}}
  & \makecell[c]{\footnotesize\textbf{Para-}\\\footnotesize\textbf{phrase}}
  & \makecell[c]{\footnotesize\textbf{Prompt}\\\footnotesize\textbf{Inj.}}
  & \makecell[c]{\footnotesize\textbf{Char./}\\\footnotesize\textbf{Enc.}}
  & \makecell[c]{\footnotesize\textbf{Recur-}\\\footnotesize\textbf{sive}}
  & \makecell[c]{\footnotesize\textbf{Dom./}\\\footnotesize\textbf{Fmt.}}
  & \textbf{\#Cat.} & \textbf{\#Atk.} \\
\midrule
TuringBench~\cite{uchendu2021turingbench}  & 168.6K    & 1  & \cmark & \xmark & \xmark & \xmark & \xmark & \xmark & \xmark & \xmark & \xmark & \xmark & \xmark & 0 & 0  \\
HC3~\cite{guo2023close}         & 85K    & 3  & \cmark & \xmark & \xmark & \xmark & \xmark & \xmark & \xmark & \xmark & \xmark & \xmark & \xmark & 0 & 0  \\
M4~\cite{wang2023m4}          & 24.5K  & 4  & \cmark & \xmark & \xmark & \xmark & \xmark & \xmark & \xmark & \xmark & \xmark & \xmark & \cmark & 1 & 1  \\
MAGE~\cite{li2024mage}        & 29K    & 5  & \cmark & \xmark & \xmark & \xmark & \xmark & \xmark & \cmark & \xmark & \xmark & \xmark & \cmark & 2 & 2  \\
RAID~\cite{dugan2024raid}        & 628.7K & 4  & \cmark & \xmark & \cmark & \xmark & \xmark & \xmark & \cmark & \xmark & \cmark & \xmark & \xmark & 2 & 11 \\
DetectRL~\cite{wu2025detectrl}    & 134.4K & 4  & \cmark & \cmark & \xmark & \xmark & \xmark & \xmark & \cmark & \xmark & \xmark & \xmark & \xmark & 1 & 3  \\
HART ~\cite{Bao2025DecouplingCA}       & 84K    & 4  & \cmark & \cmark & \cmark & \xmark & \xmark & \xmark & \cmark & \xmark & \xmark & \xmark & \xmark & 1 & 3  \\
ImBD ~\cite{chen2025imitate}       & 10K    & 3  & \cmark & \cmark & \cmark & \xmark & \xmark & \xmark & \cmark & \xmark & \xmark & \xmark & \cmark & 2 & 2  \\
MIRAGE~\cite{fu2025detectanyllm}      & 93.8K  & 5  & \cmark & \cmark & \cmark & \xmark & \xmark & \xmark & \cmark & \xmark & \xmark & \xmark & \cmark & 2 & 3  \\
\hdashline \\[-6pt]
\rowcolor[HTML]{FFF5F5}
\textbf{MOSAIC (Ours)} & \textbf{16K} & \textbf{5} & \cmark & \cmark & \cmark & \cmark & \cmark & \cmark & \cmark & \cmark & \cmark & \cmark & \cmark & \textbf{8} & \textbf{36} \\[2pt]
\Xhline{1.2pt}
\end{tabular}%
}
\caption{Comparison with existing MGT detection benchmarks. Our benchmark provides \textbf{36} sub-methods across \textbf{8} linguistic granularity levels with full category coverage, generated via uniform random model assignment over 6 frontier LLMs. \,\cmark\,=\,supported;\;\xmark\,=\,not supported. Comprehensive details are provided in the Appendix Sec.~\ref{sec:appendix_mosaic}}
 \label{tab:benchmark_comparison}
\end{table*}

\section{The MOSAIC Benchmark}
\label{sec:mosaic}

To systematically diagnose the vulnerabilities of existing MGTD paradigms, we introduce \textbf{MOSAIC}, a large-scale adversarial benchmark comprising 16000 human-AI text pairs. MOSAIC is generated through a rigorous pipeline emphasizing multi-granularity attacks and cross-model generation. 

\subsection{Source Data Collection and Purification}
To ensure domain diversity, we collect 36753 raw human-written texts from eight distinct sources. To obtain high-confidence human-written seed texts, we implement a stringent four-stage purification pipeline: (1) \textbf{Exact Deduplication}; (2) \textbf{Rule-Based Filtering} to remove truncated, HTML-heavy, or non-English texts; (3) \textbf{Near-Duplicate Removal} via 64-bit SimHash and Locality-Sensitive Hashing (LSH); and (4) \textbf{Multi-Dimensional Quality Scoring}, evaluating Log-TTR, punctuation, and syntax, retaining texts with scores $\ge 85$. This rigorous process yields 16000 high-quality human seed texts.

\subsection{Full-Granularity Adversarial Taxonomy}
As summarized in Tab.~\ref{tab:benchmark_comparison}, existing benchmarks cover limited attack surfaces. Even comprehensive datasets like RAID and MIRAGE encompass at most 11 attacks across 2 categories, leaving significant blind spots. In contrast, MOSAIC introduces an unprecedented taxonomy of 36 distinct adversarial prompts across 8 linguistic granularities, achieving full-spectrum coverage:

\begin{itemize}
    \setlength{\itemsep}{0pt}
    \setlength{\parskip}{0pt}
    \setlength{\parsep}{0pt}
    \item \textbf{Interleaved Human-AI Text:} Paragraph-level content replacement and sentence-level mixed interleaving attacks.
    \item \textbf{Statistical Hijacking:} Manipulating AI statistical signatures (e.g., perplexity) via controlled synonym swapping.
    \item \textbf{Multi-hop Translation Laundering:} Erasing latent AI fingerprints via cyclic multilingual translation chains.
    \item \textbf{Paraphrase \& Polish:} A continuous rewriting spectrum from light polishing to deep structural paraphrasing.
    \item \textbf{Prompt Injection \& Role-playing:} Altering default semantic styles using persona adoption and explicit anti-detection instructions.
    \item \textbf{Character \& Encoding Level:} Injecting adversarial homoglyphs and invisible zero-width characters to disrupt tokenizers.
    \item \textbf{Recursive \& Multi-Pass:} Iteratively refining generated texts across multiple models to smooth artifacts.
    \item \textbf{Domain-Specific \& Format:} Forcing highly specialized writing formats (e.g., academic hedging, internet slang).
\end{itemize}
\begin{table*}[ht]  
\centering  
\renewcommand{\arraystretch}{1.2} 
\setlength{\tabcolsep}{3pt} 
\resizebox{\textwidth}{!}{ 
\begin{tabular}{l|cccc|cccc|cccc}  
\Xhline{1.2pt}  
\multicolumn{13}{c}{\textbf{MIRAGE-DIG (Disjoint-Input Generation)}} \\ \hline  
\multirow{2}{*}{Methods} & \multicolumn{4}{c|}{Generate} & \multicolumn{4}{c|}{Polish} & \multicolumn{4}{c}{Rewrite} \\  
 & AUROC & Accuracy & MCC & TPR@5\% & AUROC & Accuracy & MCC & TPR@5\% & AUROC & Accuracy & MCC & TPR@5\% \\ \hline  
Likelihood~\cite{solaiman2019release} & 0.4936 & 0.5091 & 0.0183 & 0.0147 & 0.4653 & 0.5000 & 0.0000 & 0.0214 & 0.4337 & 0.5000 & 0.0000 & 0.0148 \\  
LogRank~\cite{automatic} & 0.4992 & 0.5128 & 0.0260 & 0.0220 & 0.4512 & 0.5000 & 0.0000 & 0.0195 & 0.4225 & 0.5000 & 0.0000 & 0.0132 \\  
Entropy~\cite{gehrmann2019gltr} & 0.6522 & 0.6150 & 0.2543 & 0.1099 & 0.5543 & 0.5417 & 0.1247 & 0.0954 & 0.5805 & 0.5566 & 0.1650 & 0.1189 \\  
RoBERTa-Base~\cite{liu2019roberta}  & 0.5523 & 0.5397 & 0.1434 & 0.1250 & 0.4859 & 0.5010 & 0.0088 & 0.0460 & 0.5020 & 0.5049 & 0.0293 & 0.0569 \\  
RoBERTa-Large~\cite{liu2019roberta}  & 0.4716 & 0.5217 & 0.0842 & 0.0871 & 0.5171 & 0.5151 & 0.0340 & 0.0633 & 0.5570 & 0.5385 & 0.0864 & 0.0895 \\  
LRR~\cite{su2023detectllm} & 0.5215 & 0.5341 & 0.0777 & 0.0701 & 0.4081 & 0.5000 & 0.0000 & 0.0200 & 0.3930 & 0.5000 & 0.0000 & 0.0188 \\  
DNA-GPT~\cite{yang2023dna}  & 0.5733 & 0.5595 & 0.1196 & 0.0776 & 0.4771 & 0.5004 & 0.0110 & 0.0309 & 0.4453 & 0.5001 & 0.0080 & 0.0251 \\  
NPR~\cite{su2023detectllm}  & 0.6120 & 0.6140 & 0.2604 & 0.0191 & 0.5071 & 0.5370 & 0.1071 & 0.0318 & 0.4710 & 0.5201 & 0.0663 & 0.0226 \\  
DetectGPT~\cite{mitchell2023detectgpt} & 0.6402 & 0.6258 & 0.2758 & 0.0275 & 0.5469 & 0.5531 & 0.1328 & 0.0355 & 0.5061 & 0.5266 & 0.0826 & 0.0283 \\  
Fast-DetectGPT~\cite{bao2023fast} & 0.7768 & 0.7234 & 0.4628 & 0.4310 & 0.5720 & 0.5570 & 0.1293 & 0.1189 & 0.5455 & 0.5432 & 0.1015 & 0.1025 \\  
ImBD~\cite{chen2025imitate} & 0.8597 & 0.7738 & 0.5497 & 0.4065 & 0.7888 & 0.7148 & 0.4300 & 0.2730 & 0.7825 & 0.7068 & 0.4139 & 0.2933 \\  
DetectAnyLLM~\cite{fu2025detectanyllm}& 0.9525 & 0.8988 & 0.7975 & 0.7770 & 0.9297 & 0.8732 & 0.7487 & 0.7756 & 0.9234 & 0.8705 & 0.7447 & 0.7778 \\ \hdashline  
\rowcolor[HTML]{FFF5F5}   
\textbf{NeuroStat} (Qwen2-0.5B)  & \textbf{0.9955} & \textbf{0.9738} & \textbf{0.9477} & \textbf{0.9851} & \textbf{0.9576} & \textbf{0.9161} & \textbf{0.8339} & \textbf{0.8818} & \textbf{0.9490} & \textbf{0.9108} & \textbf{0.8242} & \textbf{0.8710} \\  
\rowcolor[HTML]{FFF5F5}   
\textbf{NeuroStat} (Qwen3.5-0.8B) & \textbf{0.9948} & \textbf{0.9830} & \textbf{0.9661} & \textbf{0.9875} & \textbf{0.9716} & \textbf{0.9479} & \textbf{0.8979} & \textbf{0.9344} & \textbf{0.9694} & \textbf{0.9466} & \textbf{0.8951} & \textbf{0.9287} \\ \Xhline{1.2pt}  
\multicolumn{13}{c}{\textbf{MIRAGE-SIG (Shared-Input Generation)}} \\ \hline  
\multirow{2}{*}{Methods} & \multicolumn{4}{c|}{Generate} & \multicolumn{4}{c|}{Polish} & \multicolumn{4}{c}{Rewrite} \\  
 & AUROC & Accuracy & MCC & TPR@5\% & AUROC & Accuracy & MCC & TPR@5\% & AUROC & Accuracy & MCC & TPR@5\% \\ \hline  
Likelihood~\cite{solaiman2019release} & 0.4968 & 0.5207 & 0.0196 & 0.0145 & 0.4599 & 0.5002 & 0.0030 & 0.0233 & 0.4319 & 0.5000 & 0.0000 & 0.0111 \\  
LogRank~\cite{automatic} & 0.5008 & 0.5183 & 0.0182 & 0.0186 & 0.4468 & 0.5000 & 0.0000 & 0.0211 & 0.4221 & 0.5000 & 0.0000 & 0.0118 \\  
Entropy~\cite{gehrmann2019gltr} & 0.6442 & 0.6123 & 0.1592 & 0.1074 & 0.5640 & 0.5439 & 0.0516 & 0.0946 & 0.5858 & 0.5645 & 0.0918 & 0.1198 \\  
RoBERTa-Base~\cite{liu2019roberta} & 0.5368 & 0.5392 & 0.0529 & 0.1101 & 0.4741 & 0.5011 & 0.0048 & 0.0395 & 0.5099 & 0.5122 & 0.0221 & 0.0668 \\  
RoBERTa-Large~\cite{liu2019roberta} & 0.4703 & 0.5236 & 0.0417 & 0.0910 & 0.5150 & 0.5157 & 0.0283 & 0.0702 & 0.5576 & 0.5426 & 0.0405 & 0.0762 \\  
LRR~\cite{su2023detectllm} & 0.5214 & 0.5311 & 0.0314 & 0.0657 & 0.4076 & 0.5000 & 0.0000 & 0.0238 & 0.3978 & 0.5000 & 0.0000 & 0.0174 \\  
DNA-GPT~\cite{yang2023dna} & 0.5759 & 0.5647 & 0.0603 & 0.0813 & 0.4788 & 0.5001 & 0.0036 & 0.0340 & 0.4457 & 0.5002 & 0.0048 & 0.0258 \\  
NPR~\cite{su2023detectllm} & 0.6088 & 0.6170 & 0.1571 & 0.0185 & 0.5074 & 0.5277 & 0.0612 & 0.0293 & 0.4738 & 0.5204 & 0.0340 & 0.0177 \\  
DetectGPT~\cite{mitchell2023detectgpt} & 0.6353 & 0.6241 & 0.1719 & 0.0193 & 0.5434 & 0.5515 & 0.0668 & 0.0309 & 0.5079 & 0.5260 & 0.0431 & 0.0239 \\  
Fast-DetectGPT~\cite{bao2023fast} & 0.7706 & 0.7193 & 0.2078 & 0.4200 & 0.5727 & 0.5619 & 0.0607 & 0.1238 & 0.5480 & 0.5495 & 0.0525 & 0.1097 \\  
ImBD~\cite{chen2025imitate} & 0.8612 & 0.7791 & 0.5599 & 0.4183 & 0.7951 & 0.7199 & 0.4451 & 0.3036 & 0.7694 & 0.6920 & 0.3936 & 0.2868 \\  
DetectAnyLLM~\cite{fu2025detectanyllm} & 0.9526 & 0.9059 & 0.8119 & 0.7722 & 0.9316 & 0.8740 & 0.7483 & 0.7779 & 0.9158 & 0.8643 & 0.7320 & 0.7574 \\ \hdashline  
\rowcolor[HTML]{FFF5F5}   
\textbf{NeuroStat} (Qwen2-0.5B)  & \textbf{0.9937} & \textbf{0.9705} & \textbf{0.9410} & \textbf{0.9792} & \textbf{0.9560} & \textbf{0.9155} & \textbf{0.8327} & \textbf{0.8795} & \textbf{0.9497} & \textbf{0.9131} & \textbf{0.8290} & \textbf{0.8758} \\  
\rowcolor[HTML]{FFF5F5}   
\textbf{NeuroStat} (Qwen3.5-0.8B) & \textbf{0.9954} & \textbf{0.9839} & \textbf{0.9678} & \textbf{0.9871} & \textbf{0.9739} & \textbf{0.9527} & \textbf{0.9066} & \textbf{0.9381} & \textbf{0.9665} & \textbf{0.9456} & \textbf{0.8927} & \textbf{0.9290} \\ \Xhline{1.2pt}  
\end{tabular}  
}  
\caption{Experimental results across three tasks (Generate, Polish, Rewrite). Metrics include AUROC, Accuracy, MCC, and TPR@5\%. Best results are highlighted in bold.}  
\label{tab_mirage_results}  
\end{table*}  

\subsection{Multi-Model Cross-Generation}
Previous datasets over-rely on a single model family, causing TB detectors to overfit to specific ``semantic fingerprints''. To mitigate generator-specific fingerprint bias, MOSAIC employs a random cross-allocation mechanism across six frontier LLMs: GLM-5.0, GPT-5.4, Gemini-3.1-Pro, MiniMax-M2.7, Claude-Opus-4-6, and Kimi-K2.5. Specifically, the 16000 seed texts are stratified by length. Within each stratum, the 36 adversarial prompts are uniformly assigned, and each prompt-seed pair is allocated to one of the six LLMs. This model-agnostic design forces detectors evaluated on MOSAIC to learn the intrinsic boundary between human and AI texts, rather than merely memorizing the stylistic quirks of a specific generator.

\begin{table*}[htbp]  
\centering  
\small  
\renewcommand{\arraystretch}{1.3} 
\setlength{\tabcolsep}{2pt} 
\resizebox{\textwidth}{!}{ 
\begin{tabular}{l|cccccccc|cc}  
\Xhline{1.2pt}  
\multicolumn{11}{c}{\textbf{Robustness Evaluation on the MOSAIC Adversarial Benchmark}} \\ \hline  
\multirow{2}{*}{Methods} & \multicolumn{8}{c|}{\textbf{AUROC across 8 Adversarial Attack Categories}} & \multicolumn{2}{c}{\textbf{Overall Performance (\%)}} \\  
 & Interleaved & Stat. Hijack & Translation & Paraphrase & Prompt Inj. & Char./Enc. & Recursive & Dom./Fmt. & \textbf{AUROC} & \textbf{TPR@5\%} \\ \hline  
Likelihood \cite{solaiman2019release} & 49.8 & 30.6 & 62.8 & 47.0 & 47.8 & 34.1 & 46.4 & 50.0 & 46.1 & 3.5 \\  
LogRank \cite{automatic} & 50.3 & 69.9 & 37.9 & 53.4 & 53.1 & 65.5 & 54.5 & 51.2 & 54.5 & 9.3 \\  
Entropy \cite{gehrmann2019gltr} & 48.2 & 65.1 & 40.8 & 50.3 & 50.1 & 60.7 & 49.9 & 49.5 & 51.8 & 6.9 \\  
RoBERTa-Base \cite{liu2019roberta}  & 47.6 & 46.8 & 55.6 & 50.0 & 45.2 & 46.4 & 47.9 & 43.9 & 47.9 & 3.0 \\  
RoBERTa-Large \cite{liu2019roberta}  & 43.6 & 47.1 & 58.0 & 44.8 & 36.6 & 44.0 & 38.7 & 36.9 & 43.7 & 4.5 \\  
LRR \cite{su2023detectllm} & 49.7 & 30.1 & 57.9 & 45.7 & 44.7 & 37.6 & 42.6 & 45.9 & 44.3 & 3.1 \\  
DNA-GPT \cite{yang2023dna}  & 47.4 & 31.8 & 60.5 & 45.0 & 48.8 & 39.5 & 44.8 & 54.3 & 46.5 & 4.6 \\  
Fast-DetectGPT \cite{bao2023fast} & 45.5 & 34.5 & 61.0 & 44.4 & 45.7 & 44.9 & 40.1 & 54.8 & 46.4 & 6.8 \\  
Binoculars \cite{hans2024spotting}& 48.1 & 34.7 & 61.9 & 47.2 & 52.2 & 40.6 & 45.3 & 58.4 & 48.6 & 4.4 \\
TextFluoroscopy \cite{textfluor} & 56.0 & 54.6 & 58.0 & 60.1 & 58.9 & 46.5 & 60.9 & 68.1 & 57.9 & 14.4 \\
ImBD \cite{chen2025imitate}  & 68.7 & 61.9 & \underline{73.1} & 70.4 & 77.7 & 52.3 & 79.8 & 89.8 & 71.7 & 22.5 \\  
DetectAnyLLM \cite{fu2025detectanyllm}  & \underline{71.4} & \textbf{81.2} & 70.2 & \underline{74.9} & 82.8 & 52.7 & 88.2 & 95.1 & 77.1 & 40.5 \\ \hdashline  
\rowcolor[HTML]{FFF5F5}   
\textbf{NeuroStat} (Qwen2-0.5B) & \textbf{76.9} & \underline{80.5} & \textbf{77.3} & \textbf{77.3} & \textbf{89.9} & \underline{71.2} & \textbf{95.5} & \textbf{98.9} & \textbf{83.4} & \underline{57.7} \\  
\rowcolor[HTML]{FFF5F5}   
\textbf{NeuroStat} (Qwen3.5-0.8B)  & 67.8 & 77.1 & 66.5 & 68.1 & \underline{86.2} & \textbf{72.4} & \underline{89.3} & \underline{97.4} & \underline{78.1} & \textbf{58.1} \\ \Xhline{1.2pt}  
\end{tabular}  
}  
\caption{Adversarial robustness evaluation on the \textbf{MOSAIC} benchmark. The left section reports the AUROC scores across all 8 fine-grained attack categories, demonstrating the detectors' resilience against specific spoofing strategies (e.g., Statistical Hijacking, Interleaved texts). The right section reports the overall detection performance. Best results are highlighted in \textbf{bold}, and second-best results are \underline{underlined}.}  
\label{tab_mosaic_results}  
\end{table*}

\section{Experiments}
\label{sec:experiments}

\subsection{Experiment Settings}
\label{sec:exp_settings}

\boldparagraph{Datasets and Tasks.}
To comprehensively evaluate NeuroStat, we conduct experiments on three benchmarks, each targeting a different capability of the detector. \textbf{MIRAGE} evaluates \emph{generalization} across three tasks (\textsc{Generate}, \textsc{Polish}, and \textsc{Rewrite}) under both Disjoint-Input (DIG) and Shared-Input (SIG) settings. \textbf{MOSAIC}, our proposed benchmark, evaluates \emph{adversarial robustness} through 36 fine-grained attacks spanning 8 categories. \textbf{ImBD Test Set} evaluates \emph{standard detection} on human-written texts and their machine-polished counterparts across three domains. The corresponding results are provided in the Appendix Sec.~\ref{sec:appendix_imbd}.

\boldparagraph{Baselines.}
We compare NeuroStat against 11 state-of-the-art MGTD methods, including \textbf{Training-Free (TF) methods}: Likelihood~\cite{solaiman2019release}, LogRank~\cite{automatic}, Entropy~\cite{gehrmann2019gltr}, LRR~\cite{su2023detectllm}, DNA-GPT~\cite{yang2023dna}, NPR~\cite{su2023detectllm}, DetectGPT~\cite{mitchell2023detectgpt}, and Fast-DetectGPT~\cite{bao2023fast}; and \textbf{Training-Based (TB) methods}: RoBERTa-Base/Large~\cite{liu2019roberta}, ImBD~\cite{chen2025imitate} and DetectAnyLLM~\cite{fu2025detectanyllm}. On the adversarial MOSAIC benchmark (Tab.~\ref{tab_mosaic_results}), we additionally report Binoculars~\cite{hans2024spotting} and TextFluoroscopy~\cite{textfluor} as paraphrase-robust baselines, while omitting NPR and DetectGPT, whose perturbation-based scoring degrades to near-random performance under heavy adversarial rewriting. All TF baselines share the same Qwen2-0.5B reference model, and all TB baselines are retrained on the identical 500-pair split for a fair comparison.

\boldparagraph{Implementation Details.}
\textbf{(1) Base Model:} We instantiate NeuroStat using two lightweight CausalLM backbones: Qwen2-0.5B and Qwen3.5-0.8B. 
\textbf{(2) Training Paradigm:} To ensure a strictly fair comparison, NeuroStat is trained on only 500 pairs of human-written and machine-polished texts, identical to the ImBD training split. Diverging from previous methods that freeze the backbone or rely on parameter-efficient fine-tuning (e.g., LoRA), the entire NeuroStat framework, including the CausalLM backbone and the dual-branch modules, is trained end-to-end. This 500-pair budget is chosen purely as a fairness constraint rather than an assumed sufficiency threshold, and a training-size sensitivity study is provided in Appendix Sec.~\ref{sec:appendix_sensitivity}.

\subsection{Main Results}
\label{sec:main_results}
\boldparagraph{Generalization on MIRAGE Benchmark.}
Tab.~\ref{tab_mirage_results} presents the comprehensive results on the MIRAGE benchmark. Existing baselines exhibit significant performance degradation when faced with diverse domains and tasks. In contrast, NeuroStat demonstrates exceptional generalization capabilities. Under the challenging MIRAGE-SIG setting, NeuroStat (Qwen3.5-0.8B) achieves an AUROC of \textbf{0.9954}, \textbf{0.9739}, and \textbf{0.9665} across the three tasks. This strong performance with only 500 training samples highlights the superiority of our dual-branch feature extraction in capturing universal AI artifacts across diverse generative distributions.

\boldparagraph{Robustness on MOSAIC Benchmark.}
NeuroStat's superiority is most pronounced under adversarial conditions (Tab.~\ref{tab_mosaic_results}). As hypothesized, existing paradigms suffer catastrophic failures against targeted attacks. TF methods like Fast-DetectGPT plummet under \textit{Statistical Hijacking} (34.5) and \textit{Interleaved} texts (45.5). Conversely, TB methods like ImBD struggle against \textit{Translation Laundering} (73.1) and \textit{Prompt Injection} (77.7), which successfully mask semantic fingerprints. Particularly under the most challenging \textit{Character \& Encoding} attacks, which disrupt tokenizers and degrade DetectAnyLLM to near-random guessing (52.7), NeuroStat maintains a robust \textbf{72.4}. This resilience stems from a structurally asymmetric attack surface faced by attackers, since altering semantics exposes probability anomalies to the TF branch while manipulating statistics disrupts semantic coherence caught by the TB branch. Consequently, NeuroStat (Qwen2-0.5B) achieves an overall AUROC of \textbf{83.4}, and NeuroStat (Qwen3.5-0.8B) achieves a TPR@5\% of \textbf{58.1}, an absolute improvement of up to \textbf{17.6} points in this metric over the previous SOTA, with the best AUROC and the best TPR@5\% attained by different model variants.

\subsection{Qualitative Analysis}
To further understand the behavior of NeuroStat, we provide two qualitative analyses.

\begin{figure*}[t]
  \centering
  \includegraphics[width=1.0\linewidth]{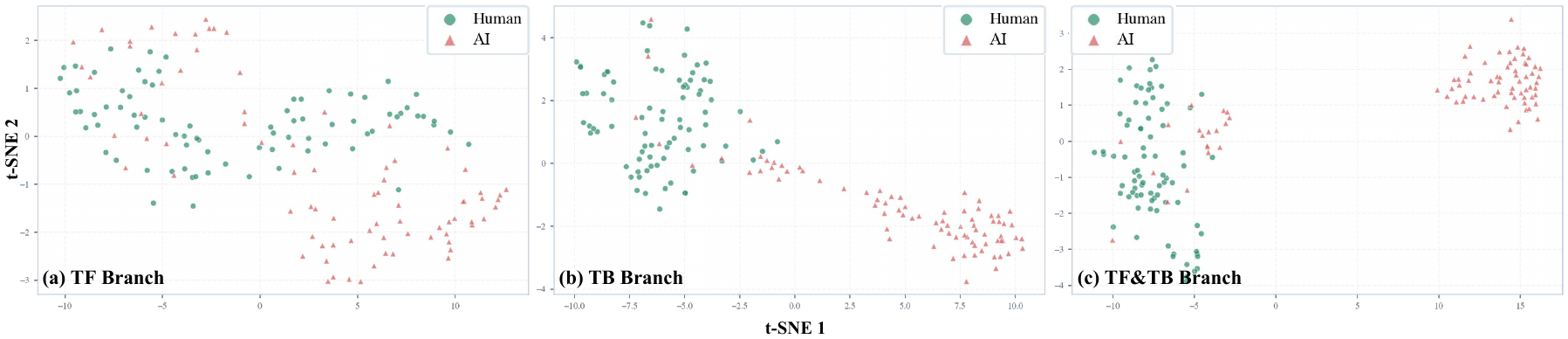}
  \caption{t-SNE visualization of TF, TB, and fused representations. The fusion space shows better class separation.}
  \label{fig:tsne}
\end{figure*}

\boldparagraph{Representation Visualization.}
We visualize the t-SNE representations on the MIRAGE Polish subset. While the TF branch exhibits entangled distributions (Fig.~\ref{fig:tsne}a) and the TB branch yields tighter but partially overlapping clusters (Fig.~\ref{fig:tsne}b), their fusion maps the classes into highly compact, linearly separable clusters with a large margin (Fig.~\ref{fig:tsne}c). This visually confirms that our dual-branch architecture successfully synergizes complementary signals, while $\mathcal{L}_{\text{SupCon}}$ effectively enforces discriminative boundaries in the joint latent space.

\begin{figure}
  \centering
  \includegraphics[width=\columnwidth]{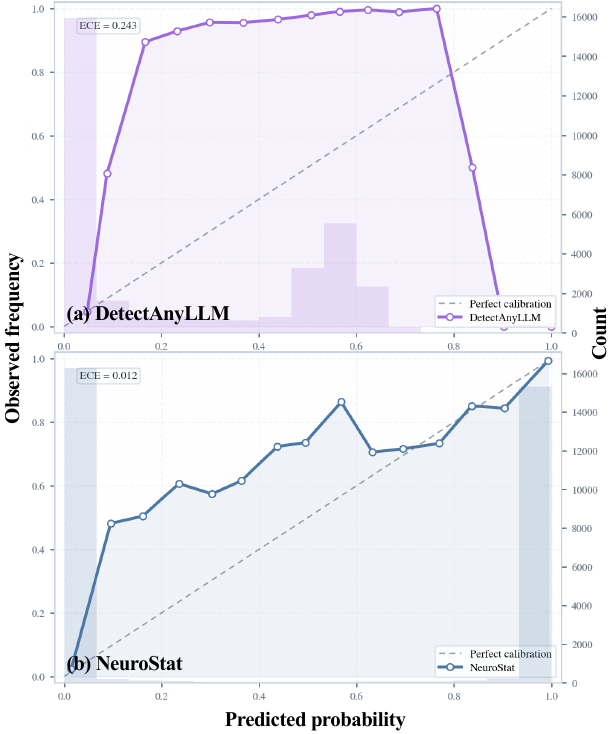}
  \vspace{-8mm}
  \caption{Reliability diagrams of NeuroStat vs.\ DetectAnyLLM. Lower ECE indicates better calibration.}
  \label{fig:calibration}
\end{figure}

\boldparagraph{Confidence Calibration.}
We compare the reliability diagrams of NeuroStat and DetectAnyLLM on the same test set (Fig.~\ref{fig:calibration}).
NeuroStat achieves a notably lower Expected Calibration Error (ECE), with predicted confidence closely tracking actual accuracy across all bins.
This suggests that the supervised contrastive and orthogonal regularization objectives not only improve discrimination but also yield better-calibrated predictions.

\section{Ablation Studies}
\label{sec:ablation}
To rigorously validate the contribution of each component in NeuroStat, we conduct ablation studies on the MIRAGE DIG benchmark using the Qwen2-0.5B backbone. We report AUROC across three text manipulation types.

\subsection{Component Ablation}
We investigate the necessity of each architectural component by progressively integrating modules into the baseline. Results are presented in Tab.~\ref{tab:ablation_component}.
\begin{table}[h]
\centering
\resizebox{\linewidth}{!}{
\begin{tabular}{lcccccc}
\toprule
\multirow{2}{*}{\textbf{Configuration}} & \multirow{2}{*}{\textbf{TF}} & \multirow{2}{*}{\textbf{TB}} & \multirow{2}{*}{\textbf{MSRM}} & \multicolumn{3}{c}{\textbf{MIRAGE DIG (AUROC)}} \\
\cmidrule(lr){5-7}
 &  &  &  & Polish & Rewrite & Generate \\
\midrule
TB-Only (Seq.\ Cls.) & $\times$ & \checkmark & $\times$ & 0.8990 & 0.8918 & 0.9787 \\
TF + TB (Concat) & \checkmark & \checkmark & $\times$ & 0.9280 & 0.9145 & 0.9934 \\
\midrule
\textbf{NeuroStat (Full)} & \checkmark & \checkmark & \checkmark & \textbf{0.9576} & \textbf{0.9490} & \textbf{0.9955} \\
\bottomrule
\end{tabular}
}
\caption{Ablation on core architectural components. ``TB-Only'' is a standard sequence classification model.}
\label{tab:ablation_component}
\end{table}
As shown in Tab.~\ref{tab:ablation_component}, a standard sequence classifier performs reasonably well on the easier \textit{Generate} subset but struggles significantly on complex manipulations. While naively concatenating the TF branch provides consistent improvements, replacing it with our MSRM mechanism yields substantial leaps on these challenging subsets. This confirms our hypothesis, rather than simple feature concatenation, MSRM's residual gating effectively amplifies local probability anomalies conditioned on global uncertainty, making it critical for detecting sophisticated text revisions.

\subsection{Ablation on Auxiliary Losses}
We evaluate the impact of the two auxiliary losses, Supervised Contrastive Loss ($\mathcal{L}_\text{SupCon}$) and Orthogonal Penalty ($\mathcal{L}_\text{Orth}$), by removing them individually and jointly.
\begin{table}[h]
\centering
\resizebox{\linewidth}{!}{
\begin{tabular}{lcccccc}
\toprule
\multirow{2}{*}{\textbf{Configuration}} & \multirow{2}{*}{$\mathcal{L}_{CE}$} & \multirow{2}{*}{$\mathcal{L}_\mathrm{SupCon}$} & \multirow{2}{*}{$\mathcal{L}_\mathrm{Orth}$} & \multicolumn{3}{c}{\textbf{MIRAGE DIG (AUROC)}}  \\
\cmidrule(lr){5-7}
 & & & & Polish & Rewrite & Generate  \\
\midrule
CE Only & \checkmark & $\times$ & $\times$ & 0.8859 & 0.8798 & 0.9850 \\
+ $\mathcal{L}_\mathrm{SupCon}$ & \checkmark & \checkmark & $\times$ & 0.9179 & 0.9124 & 0.9922  \\
+ $\mathcal{L}_\mathrm{Orth}$ & \checkmark & $\times$ & \checkmark & 0.9129 & 0.9051 & 0.9865 \\
\midrule
\textbf{NeuroStat (Full)} & \checkmark & \checkmark & \checkmark & \textbf{0.9576} & \textbf{0.9490} & \textbf{0.9955}  \\
\bottomrule
\end{tabular}
}
\caption{Impact of auxiliary losses. All variants use the full architecture, differing only in the loss configuration. }
\label{tab:ablation_loss}
\end{table}
Tab.~\ref{tab:ablation_loss} demonstrates the necessity of our auxiliary objectives. Training solely with Cross-Entropy yields suboptimal results. Individually, $\mathcal{L}_\text{SupCon}$ improves performance by enforcing a larger discriminative margin, while $\mathcal{L}_\text{Orth}$ prevents representational redundancy between the TF and TB branches. Crucially, their combination produces a synergistic effect that far exceeds their individual gains. This confirms that forcing the branches to capture orthogonal features perfectly complements the maximization of joint discriminative power. Two further ablations, attributing the gains to our architecture rather than to added fine-tuning capacity, and breaking down component and fusion mechanism choices directly on MOSAIC, are provided in Appendix Sec.~\ref{sec:appendix_ablation}.

\section{Conclusion}
We present NeuroStat, a novel end-to-end framework that bridges the Training-Free (TF) and Training-Based (TB) paradigms for robust machine-generated text detection. By preserving uncompressed probabilistic trajectories and employing the Macro-State Residual Modulation (MSRM) mechanism, NeuroStat effectively captures local probability anomalies while avoiding the lossy compression of traditional global scalars. Furthermore, we introduce MOSAIC, a comprehensive adversarial benchmark featuring 36 fine-grained attacks across 8 categories and 6 frontier LLMs. Extensive evaluations demonstrate that NeuroStat achieves state-of-the-art robustness against severe adversarial spoofing and statistical hijacking. Ultimately, this work establishes a new paradigm for synergizing statistical mechanics and semantic artifacts in AIGC detection.

\section*{Limitations}
While NeuroStat demonstrates exceptional robustness in adversarial scenarios, several limitations warrant future investigation. First, our current evaluation is constrained to English-language texts across specific domains (e.g., news, creative writing, and biomedical abstracts). The framework's effectiveness on multilingual corpora (particularly low-resource languages) or highly specialized domains such as code generation and legal documents remains unexplored. Future work should extend the MOSAIC benchmark and NeuroStat's evaluation to diverse cross-lingual and cross-domain settings.

Second, the Probabilistic Trajectories (TF) branch relies on extracting token-level log-probabilities from a CausalLM backbone. While this is straightforward for open-weight models, many closed-source commercial APIs do not expose their full log-probability distributions. In such black-box scenarios, our framework relies on a local open-source surrogate model to approximate the probabilistic dynamics of the target proprietary generator. Notably, our main MOSAIC evaluation already constitutes such a surrogate setting, since a single Qwen2-0.5B backbone is used to score text produced by six generators, four of which are closed-source commercial systems. As detailed in Appendix Sec.~\ref{sec:appendix_surrogate}, the resulting gap between open and closed generators is only 3 to 7 AUROC points, and NeuroStat still surpasses the strongest baseline on every closed-source target.

Finally, the training and inference processes of NeuroStat incur additional memory overhead compared to standard sequence classifiers, primarily due to the storage and processing of the full-vocabulary logit matrix. As measured in Appendix Sec.~\ref{sec:appendix_cost}, this translates to roughly 10\% higher latency and memory than the strongest end-to-end baseline DetectAnyLLM. Future work could investigate vocabulary truncation or top-$k$ logit approximation strategies to further reduce this footprint for large-scale deployments.

\section*{Ethics Statement}
This work utilizes publicly available datasets for human-written texts, including XSum, PubMedQA, WritingPrompts, and the MIRAGE benchmark. These datasets consist of news articles, scientific abstracts, and creative stories that do not contain Personally Identifiable Information (PII) or potentially harmful material. All datasets are used in strict accordance with their original licenses and intended research purposes. Furthermore, the adversarial prompts designed for the MOSAIC benchmark are strictly formulated to test statistical and semantic robustness (e.g., paraphrasing, synonym swapping, formatting) and are explicitly constrained from eliciting toxic, hateful, or offensive content from the language models.

We employ both open-source models (e.g., Qwen2, Qwen3.5) and commercial APIs to construct the MOSAIC benchmark and train our detectors. All models and APIs are accessed and utilized in full compliance with their respective terms of service and usage policies, which permit research on AI safety and detection.

Regarding broader impacts, we acknowledge the dual-use nature of Machine-Generated Text Detection (MGTD) technologies. While NeuroStat is designed to mitigate risks associated with academic dishonesty and misinformation, no detector is infallible. False positives (misclassifying human text as AI-generated) can lead to unjust punitive actions, particularly against non-native English writers whose linguistic patterns may inadvertently trigger detection thresholds. Therefore, we strongly emphasize that NeuroStat is intended strictly as a research tool to advance the understanding of LLM artifacts. It should serve as an assistive signal rather than the sole, definitive evidence for disciplinary or punitive decisions in real-world applications.

Finally, during the preparation of this manuscript, we utilized AI assistants (e.g., ChatGPT) solely for language polishing and improving the readability of the text. All core ideas, experimental designs, and data analyses are the original work of the authors, who take full responsibility for the content.



\bibliography{custom}

\appendix
\section*{Appendix}

\subsection*{Content}

This Appendix contains the following parts:
\begin{itemize}
    \item \textbf{Implementation Details}. We provide the comprehensive training configurations, optimization hyperparameters, and hardware specifications used for NeuroStat.
    \item \textbf{The MOSAIC Benchmark: Detailed Construction}. We detail the multi-stage data purification pipeline, the length-stratified generation protocol, and the quality validation statistics employed to construct the MOSAIC benchmark.
    \item \textbf{Evaluation on Standard Non-Adversarial Scenarios}. We evaluate NeuroStat on standard polished text benchmarks across multiple domains, demonstrating its state-of-the-art performance.
    \item \textbf{Additional Ablation Studies}. We attribute the observed gains to our architecture rather than to added fine-tuning capacity, and further ablate components and fusion mechanisms directly on MOSAIC.
    \item \textbf{Training-Sample Sensitivity}. We study how NeuroStat's performance scales with the number of training pairs, and justify the 500-pair training budget.
    \item \textbf{Closed-Source Deployment via Surrogate Backbones}. We quantify the performance gap when the CausalLM backbone acts as an open-source surrogate for closed-source target generators.
    \item \textbf{Computational Cost Analysis}. We report measured latency, memory, and throughput of NeuroStat against all baselines.
\end{itemize}

\vspace{2mm}
\noindent\rule{\linewidth}{0.4pt}
\vspace{4pt}

\section{Implementation Details}
\label{sec:appendix_impl}

To ensure full reproducibility, we detail the training configurations and hyperparameters used for NeuroStat. 

\boldparagraph{Training Configurations.} 
We instantiate NeuroStat using two lightweight CausalLM backbones: Qwen2-0.5B and Qwen3.5-0.8B. The entire framework, including the CausalLM backbone, the 1D-CNN, the MSRM module, and the fusion MLP, is trained end-to-end without freezing any components or using parameter-efficient fine-tuning methods (e.g., LoRA).

\boldparagraph{Hyperparameters.} 
The model is optimized using the AdamW optimizer with a learning rate of $2 \times 10^{-5}$ and a weight decay of 0.01. We apply a linear learning rate warmup for the first 50 steps, followed by a linear decay schedule. The training is conducted with a per-device batch size of 8 and a maximum sequence length of 512 tokens. All models undergo training for 5 epochs to ensure full convergence. 

\boldparagraph{Hardware and Reproducibility.} 
To accelerate training and reduce memory footprint, we utilize automatic mixed precision (bfloat16). All experiments are conducted on a single NVIDIA H20 GPU. The random seed is fixed to 42 across all experiments to ensure deterministic reproducibility.

\section{The MOSAIC Benchmark: Detailed Construction}
\label{sec:appendix_mosaic}

While Sec.~\ref{sec:mosaic} provides a high-level overview of the MOSAIC benchmark, this section details the rigorous data collection, purification pipeline, and multi-model generation protocols used to construct the dataset.

\begin{figure*}[htbp]
   \centering
   \includegraphics[width=1.0\linewidth]{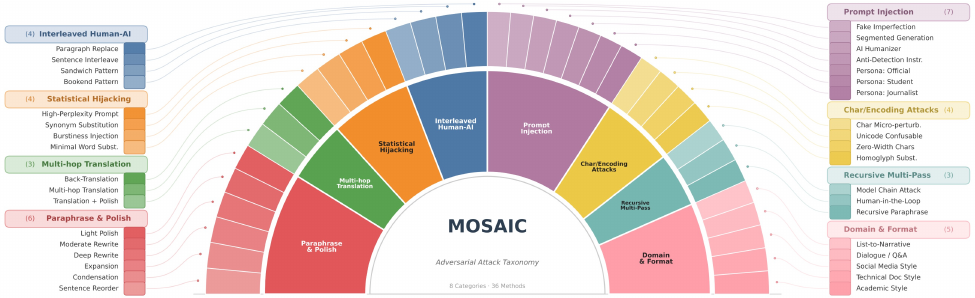}
   \caption{\textbf{The full-granularity adversarial attack taxonomy of the MOSAIC benchmark.} The sunburst chart illustrates the hierarchical structure of our dataset, encompassing 8 primary attack categories (inner circle) and 36 fine-grained sub-methods (outer circle). This comprehensive spectrum ensures a rigorous evaluation of detectors against both statistical hijacking and semantic obfuscation.}
   \label{fig:mosaic_taxonomy}
\end{figure*}

\subsection{Source Data Collection}
To ensure broad domain coverage, we aggregated 36,753 raw human-written texts from 7 distinct sources. As detailed in Tab.~\ref{tab:mosaic_sources}, these sources encompass diverse domains including news, biomedical abstracts, creative writing, and Wikipedia contexts.

\begin{table}[h]
\centering
\small
\resizebox{\linewidth}{!}{
\begin{tabular}{llc}
\toprule
\textbf{Source Dataset} & \textbf{Domain} & \textbf{Raw Samples} \\
\midrule
MIRAGE-BENCH DIG & Multi-domain & 16,411 \\
MIRAGE-BENCH SIG & Multi-domain & 16,388 \\
AI Detection 3000 & Mixed & 3,354 \\
PubMed & Biomedical & 150 \\
SQuAD & Wikipedia/QA & 150 \\
Writing Prompts & Creative Writing & 150 \\
XSum & News & 150 \\
\midrule
\textbf{Total Raw Samples} & & \textbf{36,753} \\
\bottomrule
\end{tabular}
}
\caption{Source datasets used for collecting human seed texts.}
\label{tab:mosaic_sources}
\end{table}

\subsection{Four-Stage Purification Pipeline}
To obtain high-confidence and high-quality human seed texts, we implemented a stringent four-stage cascaded filtering pipeline, reducing the raw 36,753 samples to 16,000 high-quality seeds:

\boldparagraph{Stage 1: Exact Deduplication.} We performed global exact string matching across all 7 sources, removing 15,472 duplicate samples ($-42.1\%$).

\boldparagraph{Stage 2: Rule-Based Quality Filtering.} We applied 6 heuristic rules to filter out low-quality texts, removing 926 samples ($-4.4\%$). Texts were discarded if they: (1) contained fewer than 30 words or 100 characters; (2) were truncated (ending with commas, hyphens, etc.); (3) were HTML-heavy (tag character ratio $> 5\%$); (4) contained excessive URLs ($> 3$); (5) had a non-standard special character ratio $> 15\%$; or (6) were non-English (ASCII alpha ratio $< 70\%$).

\boldparagraph{Stage 3: Near-Duplicate Removal.} To eliminate highly similar documents that evaded exact deduplication, we utilized 64-bit SimHash combined with a 4-band Locality-Sensitive Hashing (LSH) mechanism. Pairs with a Hamming distance $\le 5$ were flagged as near-duplicates, resulting in the removal of 1,254 samples ($-6.2\%$).

\boldparagraph{Stage 4: Multi-Dimensional Quality Scoring.} Finally, we designed a 6-dimensional quality scoring system (max score 100) to evaluate the linguistic richness of the remaining texts. The criteria included: Vocabulary Diversity via Log-TTR (30 pts), Sentence Length (ideal range 12--30 words, 20 pts), Sentence Count ($\ge 3$, 15 pts), Punctuation Regularity (ratio 2\%--10\%, 15 pts), Digit Ratio ($< 5\%$, 10 pts), and Capitalization Ratio (2\%--15\%, 10 pts). Only texts achieving a score $\ge 85.0$ were retained, removing 3,101 samples and yielding the final \textbf{16,000} high-quality human seed texts.

\subsection{Multi-Model Generation and Allocation Protocol}
To mitigate model-specific fingerprint bias, the 16,000 seed texts were rewritten using 6 frontier LLMs: GLM-5.0, GPT-5.4, Gemini-3.1-Pro, MiniMax-M2.7, Claude-Opus-4-6, and Kimi-K2.5. 

\boldparagraph{Length-Stratified Balanced Allocation.} 
To ensure uniform distribution across text lengths and attack types, the seed texts were first stratified into three length tiers: Short (183--500 chars, 2.5\%), Medium (501--1,000 chars, 61.9\%), and Long (1,001--1,991 chars, 35.6\%). Within each stratum, the 36 adversarial prompts were strictly and uniformly assigned. Each prompt-seed pair was then randomly allocated to one of the 6 LLMs. 

\boldparagraph{Generation Hyperparameters.} 
All prompts included a unified \texttt{global\_instruction} forcing the models to output only the rewritten text without any meta-annotations. For GLM-5.0, we used the officially recommended parameters (\texttt{temperature = 1.0}, \texttt{top\_p = 0.95}). For all other models, default generation parameters were utilized. This rigorous protocol resulted in exactly \textbf{16,000 human-AI text pairs}, ensuring a perfectly balanced and unbiased adversarial benchmark.

\subsection{Sample Distribution and Attack Validity}
\label{sec:appendix_mosaic_stats}
Each of the 16,000 human seeds is paired with one of the 36 attack sub-methods and one of the 6 frontier LLMs through length-stratified sampling rather than a full cross-product, yielding roughly 2,667 pairs per generator and a per-category count ranging from 1,777 to 2,667 pairs.

To validate rewrite quality, we audit a stratified 2,000-pair subsample. The attack success rate, defined as fooling at least one SOTA detector, ranges from 42.3\% to 78.2\% across sub-methods. Semantic preservation is measured by BERTScore-F1 (0.751 to 0.982) and GPT-4o ratings (4.0 to 4.9 out of 5). Three human annotators on 300 sampled pairs yield a Fleiss kappa of 0.71, confirming that at least 92\% of the rewrites remain fluent and topically faithful. The 36 sub-methods are also linguistically distinct rather than redundant, with a mean pairwise token-edit Jensen-Shannon divergence of 0.38 across sub-methods versus 0.07 within a sub-method.

Since Character and Encoding attacks mainly perturb tokenizer inputs, we further report a post-normalization evaluation (NFKC plus zero-width stripping). DetectAnyLLM recovers to 61.4 AUROC and NeuroStat to 74.9, so the advantage persists even after controlling for pure Unicode noise.

\section{Evaluation on Standard Non-Adversarial Scenarios}
\label{sec:appendix_imbd}
While the main text focuses on the generalization (MIRAGE) and adversarial robustness (MOSAIC) of NeuroStat, we also evaluate its performance in standard, non-adversarial scenarios. For this purpose, we utilize the ImBD test set~\cite{chen2025imitate}, a standard benchmark containing human-written texts and their machine-polished counterparts (via GPT-3.5 and GPT-4o) across three domains (XSum, WritingPrompts, PubMedQA).

\begin{table*}[ht]  
\centering  
\small 
\renewcommand{\arraystretch}{1.15} 
\resizebox{\textwidth}{!}{  
\begin{tabular}{l|ccc|ccc}  
\Xhline{1.2pt} 
\multicolumn{7}{c}{\textbf{ImBD Test Dataset (GPT-3.5 \& GPT-4o polished)}} \\ \hline  
\multirow{2}{*}{Methods} & \multicolumn{3}{c|}{GPT-3.5} & \multicolumn{3}{c}{GPT-4o} \\  
 & XSum  & Writing  & PubMed  & XSum  & Writing & PubMed  \\ \hline  
Likelihood \cite{solaiman2019release} & 0.4982 & 0.8788 & 0.5528 & 0.4396 & 0.8077 & 0.4596 \\  
LogRank \cite{automatic} & 0.4711 & 0.8496 & 0.5597 & 0.4002 & 0.7694 & 0.4472 \\  
Entropy \cite{gehrmann2019gltr} & 0.6742 & 0.3021 & 0.5662 & 0.6122 & 0.2802 & 0.5899 \\  
RoBERTa-Base \cite{liu2019roberta}  & 0.5806 & 0.7225 & 0.4370 & 0.4921 & 0.4774 & 0.2496 \\  
RoBERTa-Large \cite{liu2019roberta}  & 0.6391 & 0.7236 & 0.4848 & 0.4782 & 0.4708 & 0.3089 \\  
LRR \cite{su2023detectllm} & 0.4016 & 0.7203 & 0.5629 & 0.3095 & 0.6214 & 0.4710 \\  
DNA-GPT \cite{yang2023dna}  & 0.5338 & 0.8439 & 0.3333 & 0.4974 & 0.7478 & 0.3151 \\  
NPR \cite{su2023detectllm}  & 0.5659 & 0.8786 & 0.4246 & 0.5065 & 0.8444 & 0.3740 \\  
DetectGPT \cite{mitchell2023detectgpt}  & 0.6343 & 0.8793 & 0.5608 & 0.6217 & 0.8771 & 0.5612 \\  
Fast-DetectGPT \cite{bao2023fast} & 0.7312 & 0.9304 & 0.7182 & 0.6293 & 0.8324 & 0.6175 \\  
ImBD \cite{chen2025imitate}  & 0.9849 & 0.9871 & 0.8626 & 0.9486 & 0.9468 & 0.7743 \\  
DetectAnyLLM \cite{fu2025detectanyllm}  & \underline{0.9948} & 0.9688 & 0.8500 & 0.9884 & 0.9669 & 0.8528 \\  
 \hdashline  
\rowcolor[HTML]{FFF5F5}   
\textbf{NeuroStat} (Qwen2-0.5B)  & \textbf{0.9950} & \textbf{0.9932} & \underline{0.9216} & \underline{0.9943} & \underline{0.9774} & \underline{0.9301} \\  
\rowcolor[HTML]{FFF5F5}   
\textbf{NeuroStat} (Qwen3.5-0.8B) & 0.9944 & \underline{0.9929} & \textbf{0.9525} & \textbf{0.9972} & \textbf{0.9845} & \textbf{0.9679} \\  
\Xhline{1.2pt}   
\end{tabular}  
}  
\caption{Detection performance (AUROC) on the standard ImBD test dataset. Best results are highlighted in \textbf{bold}, and second-best results are \underline{underlined}.}  
\label{tab_imbd_results}
\end{table*}

\boldparagraph{Performance Analysis.}
Tab.~\ref{tab_imbd_results} presents the detection performance on the standard ImBD test set, where texts are polished without explicit adversarial intent. While recent Training-Based (TB) methods like ImBD achieve strong performance (e.g., 0.9871 AUROC on GPT-3.5 Writing), NeuroStat consistently pushes the boundary further. Notably, on the more challenging GPT-4o polished texts, NeuroStat (Qwen3.5-0.8B) achieves near-perfect AUROC scores (0.9972 on XSum, 0.9845 on Writing), outperforming all TF and TB baselines. This demonstrates that our dual-branch fusion not only excels in adversarial robustness but also maximizes sensitivity in standard, non-adversarial scenarios.

\section{Additional Ablation Studies}
\label{sec:appendix_ablation}

\subsection{Component and Fusion Ablation on MOSAIC}
\label{sec:ablation_mosaic}
The ablations in the main text are conducted on MIRAGE, so we additionally verify that each component's contribution transfers to adversarial conditions by ablating directly on MOSAIC (Qwen2-0.5B, Tab.~\ref{tab:ablation_mosaic_component}). Neither branch dominates alone, as the TB-only classifier reaches 74.9 average AUROC and the TF-only variant reaches 74.6, yet their strengths are complementary across categories. The TF branch leads on statistical and burst-driven attacks such as Statistical Hijacking, Interleaved text, Character/Encoding, and Recursive rewriting, while the TB branch leads on semantic attacks such as Translation Laundering, Paraphrase, Prompt Injection, and Domain/Format manipulation. Naive concatenation lifts every category to 78.8 on average, and replacing it with MSRM adds a further gain that peaks on Statistical Hijacking, Paraphrase, and Interleaved text, exactly the categories that motivated the MSRM design.
\begin{table}[h]
\centering
\resizebox{\linewidth}{!}{
\begin{tabular}{lccccccccc}
\toprule
\textbf{Config} & \textbf{Int.} & \textbf{S.H.} & \textbf{Trans.} & \textbf{Para.} & \textbf{P.I.} & \textbf{C/E} & \textbf{Rec.} & \textbf{D/F} & \textbf{Avg.} \\
\midrule
TB-Only & 68.0 & 70.0 & 71.0 & 70.0 & 82.0 & 60.0 & 85.0 & 93.0 & 74.9 \\
TF-Only & 70.5 & 76.0 & 64.0 & 66.0 & 77.0 & 65.5 & 88.0 & 90.0 & 74.6 \\
Concat & 71.8 & 76.2 & 73.4 & 71.5 & 84.1 & 66.7 & 90.2 & 96.1 & 78.8 \\
\textbf{+MSRM} & \textbf{76.9} & \textbf{80.5} & \textbf{77.3} & \textbf{77.3} & \textbf{89.9} & \textbf{71.2} & \textbf{95.5} & \textbf{98.9} & \textbf{83.4} \\
\bottomrule
\end{tabular}
}
\caption{Per-category AUROC on MOSAIC. Int. Interleaved, S.H. Statistical Hijacking, Trans. Translation, Para. Paraphrase, P.I. Prompt Injection, C/E Character/Encoding, Rec. Recursive, D/F Domain/Format.}
\label{tab:ablation_mosaic_component}
\end{table}

We further compare MSRM against stronger fusion alternatives in Tab.~\ref{tab:ablation_fusion}. A pure multiplicative gate sharing the same 192-parameter projection as MSRM but omitting the residual term reaches only 79.7, isolating a 3.7-point gain that comes from the residual formulation rather than added capacity. Higher-capacity modulators such as FiLM-style conditioning, SE-style gating, and cross-attention fusion all remain below MSRM despite using more parameters. Since MSRM's multiplier $(1 + \sigma(\cdot))$ is bounded below by 1, local anomaly features are always preserved and can never be suppressed toward zero, unlike gating mechanisms whose multiplier can approach zero and erase local evidence when an attacker inflates global confidence.
\begin{table}[h]
\centering
\resizebox{\linewidth}{!}{
\begin{tabular}{lcc}
\toprule
\textbf{Fusion Mechanism} & \textbf{Params} & \textbf{MOSAIC AUROC} \\
\midrule
Naive concatenation & 0 & 78.8 \\
Multiplicative gate (w/o residual) & 192 & 79.7 \\
Log-rank-only gate & 96 & 80.6 \\
Cross-attention fusion & 16,640 & 80.9 \\
Entropy-only gate & 96 & 81.3 \\
SE-style modulation & 2,128 & 81.5 \\
FiLM conditioning & 384 & 82.1 \\
\textbf{MSRM (ours)} & \textbf{192} & \textbf{83.4} \\
\bottomrule
\end{tabular}
}
\caption{Comparison of MSRM with alternative fusion mechanisms on MOSAIC. The residual formulation, not additional parameters, accounts for the gain.}
\label{tab:ablation_fusion}
\end{table}

\section{Training-Sample Sensitivity}
\label{sec:appendix_sensitivity}
The 500-pair training budget used throughout the main text is chosen to strictly match the training size of ImBD for a fair comparison, not because we assume it is sufficient in an absolute sense. To justify this choice, we train NeuroStat (Qwen2-0.5B) with varying numbers of human-AI pairs and evaluate on both MIRAGE-Polish and MOSAIC. Tab.~\ref{tab:appendix_sensitivity} reports the resulting performance curve.

\begin{table}[h]
\centering
\resizebox{\linewidth}{!}{
\begin{tabular}{lccc}
\toprule
\textbf{Train Pairs} & \textbf{MIRAGE-Polish} & \textbf{MOSAIC Avg.} & \textbf{MOSAIC TPR@5\%} \\
\midrule
100 & 0.8891 & 71.4 & 39.7 \\
250 & 0.9247 & 78.9 & 51.6 \\
500 (default) & 0.9576 & 83.4 & 57.7 \\
1,000 & 0.9631 & 84.7 & 60.5 \\
2,000 & 0.9694 & 85.9 & 63.4 \\
5,000 & 0.9718 & 86.5 & 64.9 \\
\bottomrule
\end{tabular}
}
\caption{Training-sample sensitivity of NeuroStat (Qwen2-0.5B). Performance saturates around 2,000 to 5,000 pairs.}
\label{tab:appendix_sensitivity}
\end{table}

Two observations follow from this curve. First, NeuroStat never observes any of the 36 MOSAIC attack types during training, since it is trained only on clean, non-adversarial polished ImBD pairs, so its MOSAIC robustness emerges from the architectural inductive bias rather than from exposure to attack-specific patterns. Second, NeuroStat is data-efficient, since with only 250 training pairs it already reaches 78.9 MOSAIC AUROC, close to the strongest baseline DetectAnyLLM (77.1) trained on the full 500-pair budget (Tab.~\ref{tab_mosaic_results}). Performance saturates around 2,000 to 5,000 pairs, indicating diminishing returns beyond the 500-pair setting used in the main experiments.

\begin{table}[h]
\centering
\resizebox{\linewidth}{!}{
\begin{tabular}{lccc}
\toprule
\textbf{Target Generator} & \textbf{Access} & \textbf{AUROC} & \textbf{TPR@5\%} \\
\midrule
GPT-5.4 & Closed & 79.6 & 51.2 \\
Claude-Opus-4-6 & Closed & 81.3 & 54.7 \\
Gemini-3.1-Pro & Closed & 82.7 & 56.9 \\
MiniMax-M2.7 & Closed & 84.1 & 58.8 \\
GLM-5.0 & Open & 86.2 & 62.4 \\
Kimi-K2.5 & Open & 86.5 & 62.9 \\
\midrule
Micro-average & -- & 83.4 & 57.7 \\
\bottomrule
\end{tabular}
}
\caption{Per-generator performance of NeuroStat (Qwen2-0.5B), acting as an open-source surrogate for both open and closed-source target generators.}
\label{tab:appendix_surrogate}
\end{table}

\begin{table*}[h]
\centering
\resizebox{0.8\linewidth}{!}{
\begin{tabular}{lccccc}
\toprule
\textbf{Method} & \textbf{Train. Params} & \textbf{Peak Mem} & \textbf{Latency/Sample} & \textbf{Throughput} & \textbf{Train Time} \\
\midrule
RoBERTa-Base & 125M & 3.1GB & 6.2ms & 161/s & 3min \\
Fast-DetectGPT & 0 (frozen) & 2.4GB & 22.7ms & 44/s & -- \\
ImBD (LoRA) & 4M & 4.7GB & 25.9ms & 39/s & 12min \\
DetectAnyLLM & 0.50B & 5.8GB & 28.4ms & 35/s & 21min \\
NeuroStat (Qwen2-0.5B) & 0.51B & 6.4GB & 31.1ms & 32/s & 24min \\
NeuroStat (Qwen3.5-0.8B) & 0.82B & 9.9GB & 42.6ms & 23/s & 38min \\
\bottomrule
\end{tabular}
}
\caption{Measured computational cost on a single NVIDIA H20 GPU (batch size 8, sequence length 512).}
\label{tab:appendix_cost}
\end{table*}

\section{Closed-Source Deployment via Surrogate Backbones}
\label{sec:appendix_surrogate}
NeuroStat's Probabilistic Trajectories branch requires token-level log-probabilities from a CausalLM backbone. For closed-source generators that do not expose logits, our framework relies on a local open-source model as a surrogate backbone to score text from the unseen target generator. This setting is already embedded in our main MOSAIC evaluation, since a single Qwen2-0.5B backbone scores text from six frontier LLMs, four of which are closed-source (GPT-5.4, Claude-Opus-4-6, Gemini-3.1-Pro, MiniMax-M2.7) and two open-weight (GLM-5.0, Kimi-K2.5).
Tab.~\ref{tab:appendix_surrogate} breaks down performance by target generator.

The gap between open and closed target generators is only 3 to 7 AUROC points despite the 0.5B surrogate being far smaller and architecturally distinct from GPT-5.4, Claude, or Gemini, and NeuroStat still outperforms DetectAnyLLM (77.1 micro-average) on every closed-source target. This holds because surrogate-based detection does not require reproducing the target's exact output distribution, only scoring how well a competent language model predicts the observed tokens, a principle also underlying Fast-DetectGPT and Binoculars. MSRM further stabilizes this transfer by conditioning local features on the surrogate's own global uncertainty rather than on absolute probability values.

\section{Computational Cost Analysis}
\label{sec:appendix_cost}
We measure computational cost against representative TF and TB baselines on a single NVIDIA H20 GPU (batch size 8, sequence length 512). Tab.~\ref{tab:appendix_cost} reports trainable parameters, peak memory, per-sample latency, throughput, and training time on the 500-pair budget for 5 epochs.

Relative to DetectAnyLLM, NeuroStat (Qwen2-0.5B) adds only 0.6GB of peak memory and 2.7ms of per-sample latency, roughly 10\% overhead on both metrics, mainly from the 1D-CNN branch and the MSRM module. This modest overhead yields a substantial return, since NeuroStat improves MOSAIC AUROC by 6.3 points and TPR@5\% by 17.6 points over DetectAnyLLM.

\end{document}